\documentclass{article} 
\usepackage[preprint]{colm2026_conference}
\usepackage{amsmath} 
\usepackage{microtype}
\usepackage{hyperref}
\usepackage{url}
\usepackage{booktabs}
\usepackage{graphicx}

\usepackage{lineno}

\definecolor{darkblue}{rgb}{0, 0, 0.5}
\hypersetup{colorlinks=true, citecolor=darkblue, linkcolor=darkblue, urlcolor=darkblue}

\title{Gold-Guided Programmatic Distillation for Financial\\ Reasoning over Hybrid Tables and Text}

\author{
Yun Dong \\
College of Computing \\
Georgia Institute of Technology\\
\texttt{ydong380@gatech.edu}
\And
Erica Zhao\\
Department of Management Science \& Engineering \\
Stanford University \\
\texttt{erica117@stanford.edu}
\And
Elana Chen\\
Department of Computer Science \\
Stanford University \\
\texttt{elanac25@stanford.edu}
}

\begin{document}

\ifcolmsubmission
\linenumbers
\fi

\maketitle
\lhead{}

\begin{abstract}

Financial question answering over hybrid tabular and textual data may require multi-source reasoning and precise numerical computation. While large language models (LLMs) can generate intermediate reasoning steps, natural-language rationales remain prone to arithmetic errors, making them an unreliable supervision source for distillation.

Building on programmatic distillation, we develop an approach that transfers reliable numerical reasoning from a large teacher model to a compact student using execution-verified Python programs instead of free-form textual rationales. It leverages gold derivations to guide teacher-side program synthesis and retains only programs that execute correctly and produce the gold answer, ensuring high-quality supervision. We further introduce an iterative recovery stage that revisits teacher-failed examples, enabling the student to recover and incorporate newly verified programs into training.

Experiments on TAT-QA show that our framework is highly effective for hybrid financial reasoning. Our best 7B student achieves 87.00 EM / 87.18 F1 on the test set, substantially outperforming the 72B teacher (78.46 EM) as well as traditional and strong LLM-based baselines, including TAGOP and TAT-LLM. These results demonstrate that execution-verified programmatic distillation provides an effective and extensible framework for training smaller models to perform reliable numerical reasoning.
\end{abstract}

\section{Introduction}
Financial question answering over hybrid tabular and textual data may require integrating information from multiple sources and, in many cases, performing precise numerical reasoning. While Chain-of-Thought (CoT) prompting~\cite{wei2022chain} enables large language models (LLMs) to generate intermediate reasoning steps, their final answers remain prone to arithmetic errors, making natural-language rationales an unreliable supervision source for distillation.

Program-of-Thought (PoT)~\cite{chen2022program} addresses this limitation by expressing reasoning as executable programs and delegating computation to an external interpreter. However, effectively transferring such reliable reasoning into smaller models remains challenging, as distillation quality depends critically on the correctness of intermediate traces.

In this work, we study a gold-guided programmatic distillation framework for transferring reliable numerical reasoning from a large teacher model to a compact student. The teacher synthesizes executable Python programs guided by gold derivations, and we retain only programs that produce the gold answer, resulting in a high-quality supervision signal for distillation.

Building on this framework, we introduce an iterative recovery stage that revisits examples the teacher fails to solve. By sampling candidate programs from the student and retaining only execution-verified correct solutions, we recover additional high-quality reasoning traces, thereby expanding supervision beyond the teacher’s original coverage.

\section{Related Work}

Question answering over hybrid tabular and textual contexts, such as the TAT-QA financial benchmark~\cite{zhu2021tat}, requires complex numerical reasoning and multi-source evidence aggregation. Early approaches like TAGOP relied on rigid extraction pipelines and predefined symbolic operations, which struggle to bridge the gap to human performance~\cite{zhu2021tat}. Consequently, the field has shifted toward large language models (LLMs). Approaches such as TAT-LLM~\cite{zhu2024tat} demonstrate that LLMs can successfully tackle discrete tabular-text reasoning by decomposing inference into step-wise stages, highlighting the potential of LLM-based financial reasoning.

While explicit intermediate reasoning, such as Chain-of-Thought (CoT) prompting~\cite{wei2022chain}, helps models decompose complex tasks, natural language rationales remain highly prone to arithmetic hallucinations. Program-of-Thought (PoT) addresses this by representing reasoning as executable programs and delegating computation to an external interpreter, ensuring the precise calculation required for financial QA~\cite{chen2022program, gao2023pal}.

Crucially, these reasoning capabilities can be transferred from large teacher models to smaller students. While distilling CoT rationales improves student performance~\cite{magister2023teaching, hsieh2023distilling}, the reliability of these distilled traces dictates downstream success~\cite{tian2025not}. Frameworks such as Program-aided Distillation (PaD) improve trace quality by replacing text rationales with automatically verifiable code~\cite{zhu2024pad}. Furthermore, iterative methods such as STaR~\cite{zelikman2022star} and Rejection Sampling Fine-Tuning (RFT)~\cite{yuan2023scaling} show that models can autonomously self-improve by generating, verifying, and training on multiple candidate reasoning paths.

Taken together, prior work suggests two open challenges for distilling reliable reasoning. First, distillation quality depends heavily on the quality of intermediate traces, and imperfect or natural-language traces may propagate errors. Second, while iterative generation-and-verification strategies have shown promise, an important open question is how they can be used to expand supervision beyond the teacher’s solved examples. These challenges motivate the need for a more reliable and extensible framework for distilling executable reasoning.
\section{Method}
\subsection{Overview}
\begin{figure}[t]
\begin{center}
\includegraphics[width=0.7\linewidth]{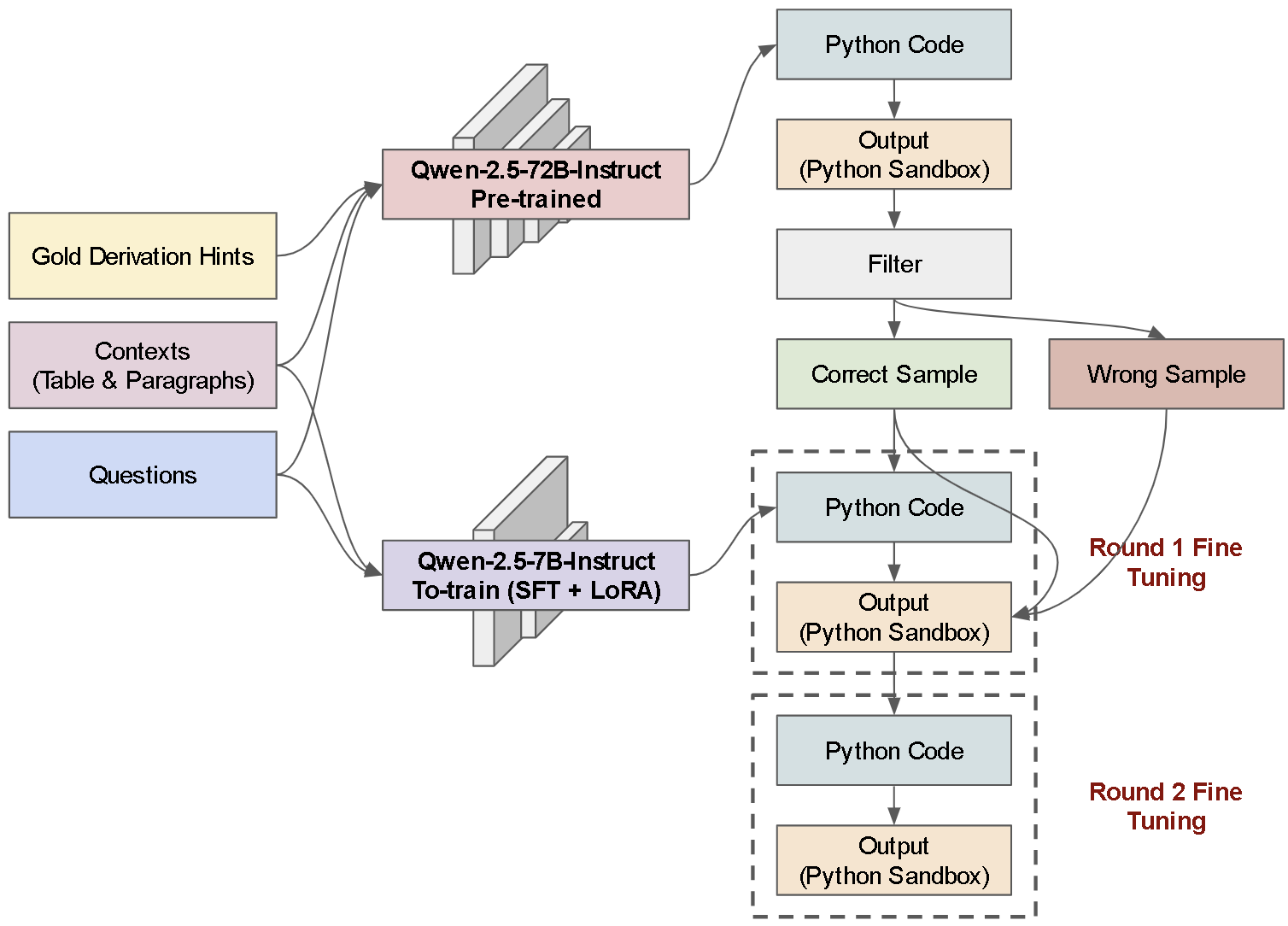}
\end{center}
\caption{Overview of our two-phase programmatic distillation and iterative fine-tuning framework.}
\label{fig:structure}
\end{figure}

We propose a programmatic distillation framework that trains a compact language model to generate executable reasoning programs for hybrid financial question answering, using the teacher model’s successful program examples as supervision. The key idea is to represent reasoning as Python programs, so that the final answer is obtained through code execution rather than the model’s internal arithmetic.\\
Our framework consists of two main stages:
(1) teacher-driven distillation, where a large model generates high-quality executable programs, and
(2) iterative self-improvement, where the student model generates new programs for teacher-failed examples and keeps only those that execute to the correct answer.
Fig. ~\ref{fig:structure} illustrates the overall pipeline of our two-phase distillation and refinement framework.
\subsection{Phase 1: Teacher-Driven Distillation}
In the first phase, we extract reasoning paths from the teacher model. Given a financial context $C$ (comprising a table $T$ and paragraphs $P$) and a query $q$, the teacher generates a Python program $\pi$. To ensure accurate execution traces and guide the teacher through complex multi-step calculations, we inject the TAT-QA ground-truth derivation $d$ into the teacher's prompt. We further apply DSPy~\cite{khattab2023dspycompilingdeclarativelanguage} for prompt optimization to format the output strictly as executable code. The generation is defined as:
$$\pi=\text{Teacher}(C,q,d)$$
To construct a high-quality distillation dataset, $D_{T}$, we execute generated programs $\pi$ in a sandboxed Python interpreter. Only programs that execute without compilation errors and yield the exact gold numerical answer ($EM = 1$) and scale are retained. The student model is then trained on $D_{T}$ via supervised fine-tuning utilizing Low-Rank Adaptation (LoRA)~\cite{hu2022lora}. 

The student optimizes the log-likelihood of generating $\pi$ conditioned solely on $C$ and $q$, without access to the ground-truth derivation $d$:
$$\mathcal{L}_{\text{Phase1}}=-\sum_{(C, q, \pi) \in D_{T}}\log P_{\text{Student}}(\pi \mid C,q)$$

\subsection{Phase 2: Iterative Rejection Sampling and Self-Improvement} 
In the first round, we retain only teacher-generated programs that execute correctly and match the gold answer, leaving teacher-failed examples in a failure set $D_{wrong}$. This constrains the student to the teacher’s successful coverage. To recover these harder cases, we introduce a second round of training inspired by Rejection Sampling Fine-Tuning (RFT) and STaR.

Starting from the Phase-1 student model, we revisit each $(C, q) \in D_{wrong}$ and sample five candidate programs ${\hat{\pi}_1, \dots, \hat{\pi}_5}$ at a higher decoding temperature ($T=0.7$):
$$\hat{\pi}_k \sim P_{\text{Student}_{v1}}(\cdot \mid C, q)$$
Each candidate is evaluated using the same sandboxed execution filter as in Phase 1. If a candidate executes successfully and produces the gold answer, it is added to a recovery set $D_{rec}$. We then perform a second round of LoRA fine-tuning on the augmented dataset $D_{\text{Total}} = D_T \cup D_{rec}$ using the same objective as in Phase 1. This stage allows the student to learn from execution-verified self-generated programs beyond the teacher’s original coverage.
\subsection{Inference} 
At inference time, the trained student model generates an executable program $\hat{\pi}$ given $(C,q)$. The program is executed by a Python interpreter to produce the final numerical answer.

This design eliminates arithmetic errors by delegating all numerical computation to deterministic execution, while the model focuses solely on reasoning.
\section{Experimental Setup}
\subsection{Dataset}
\begin{figure}[t]
\begin{center}
\includegraphics[width=\linewidth]{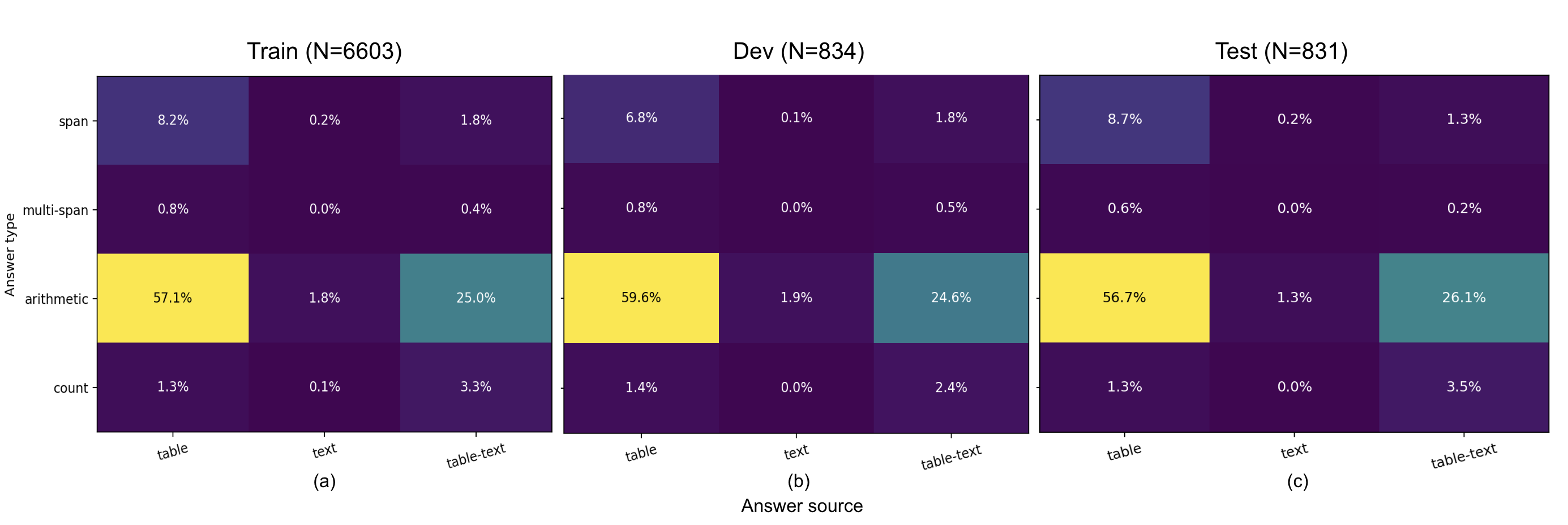} 
\end{center}
\caption{Answer type $\times$ answer source heatmaps across (a) train (b) dev (c) test. The distributions of answer types and sources remain consistent across splits, dominated by arithmetic questions answered exclusively from tables (57\%) and those requiring both tables and text (25\%).}
\label{fig:heatmaps}
\end{figure}

We utilize the TAT-QA dataset~\cite{zhu2021tat} for question answering over hybrid financial tables and text. The student model input comprises a table, associated paragraphs, and a natural-language question. A teacher setting additionally incorporates the dataset's gold derivation as an explicit reasoning hint. The models generate executable Python code to predict a structured answer and scale. Because our approach requires executable code, we exclude examples lacking derivations. All reported results are evaluated on this filtered subset for fair comparison. This yields an approximate 8:1:1 split of 6,603 training, 834 development, and 831 test examples. 
From the dataset's raw structure, we extract the 2D table array, paragraph text, and question text as inputs, reserving the gold answer and scale for evaluation. As illustrated in Fig.~\ref{fig:heatmaps}, the distributions of answer types and sources remain consistent across splits, dominated by arithmetic questions answered exclusively from tables (~57\%) and those requiring both tables and text (~25\%).

\subsection{Baselines}
We compare our method against task-specific baselines, prior LLM-based systems, and reference teacher/student models. 
\\
(1) Task-specific baseline: We include TaGOP~\cite{zhu2021tat}, a strong traditional model designed specifically for TAT-QA that performs multi-step reasoning using structured operators over table and paragraph representations. 
\\
(2) Reference models: To isolate the contribution of our programmatic distillation framework, we evaluate the untuned student model (Qwen2.5-7B-Instruct) under standard prompting without programmatic distillation. We also report the performance of the teacher model (Qwen2.5-72B-Instruct) as a reference~\cite{bai2023qwen}. 
\\
(3) Prior LLM-based baseline: We further compare against TAT-LLM~\cite{zhu2024tat}, a specialized LLM for discrete reasoning over hybrid tabular and textual financial data. For completeness, we report both its LoRA and full fine-tuning (FFT) variants. All TAT-LLM results are evaluated on the same filtered subset used in our experiments to ensure a fair comparison.
\subsection{Models and Training Setup}
We use Qwen2.5-72B-Instruct as the teacher model for data synthesis and Qwen2.5-7B-Instruct as the student model for fine-tuning. To improve the executability of teacher-generated programs, we use DSPy (MIPROv2) together with manual prompt refinement to enforce a structured output format. Specifically, the teacher is required to generate executable Python code that prints a final JSON object of the form \texttt{\{"ans": ..., "scale": ...\}}.

For Phase 1 student training, we apply Low-Rank Adaptation (LoRA) for one epoch with a learning rate of \(2 \times 10^{-4}\), rank \(r=8\), \(\alpha=16\), dropout \(=0.05\), per-device batch size \(=1\), and gradient accumulation steps \(=4\). We evaluate two LoRA configurations: an attention-only variant applied to the attention projection layers (\texttt{q\_proj}, \texttt{k\_proj}, \texttt{v\_proj}, and \texttt{o\_proj}), and an attention+MLP variant that additionally includes the feed-forward projection layers (\texttt{gate\_proj}, \texttt{up\_proj}, and \texttt{down\_proj}).

In Phase 2, for each example failed by the teacher, we sample five candidate programs from the Phase-1 student model. Programs that execute successfully and produce the correct answer are added to the training set for a second round of fine-tuning. 

Final model selection is based on dev set performance. For completeness, we also report development and test performance for intermediate models in our analysis.

We implement our training and inference pipeline using Hugging Face Transformers and related PyTorch-based libraries (e.g., PEFT~\cite{peft}), and run experiments on Modal.
\subsection{Evaluation Metrics}
We evaluate using the official TAT-QA evaluation metrics, which take both the predicted answer and scale into account. For span and multi-span questions, we normalize answers (lowercasing and punctuation removal) and compute Exact Match (EM) and token-overlap F1~\cite{rajpurkar2016squad}. For multi-span answers, F1 is computed with order-invariant Hungarian matching~\cite{kuhn1955hungarian}. For numeric answers, the evaluator converts numbers to a string after applying the predicted scale, where the supported scale labels are {\textit{""}, \textit{thousand}, \textit{million}, \textit{billion}, \textit{percent}}. If code execution or output parsing fails, we mark the prediction as \textit{null}, which receives zero EM and F1.
\section{Results}
\textbf{Main Results:} Table 1 presents the main results on the test set. We compare our method against task-specific and prior LLM-based baselines, as well as untuned teacher and student reference models. Our approach substantially outperforms all baselines and reference models. In particular, the best student model, based on Qwen2.5-7B-Instruct with Attention+MLP adaptation, achieves 87.00 EM and 87.18 F1, surpassing the 72B teacher model with hints (78.46 EM / 78.73 F1) and the strongest prior baseline, TAT-LLM (FFT), by a clear margin (82.67 EM / 83.07 F1). These results show that execution-verified programmatic distillation is highly effective for hybrid financial reasoning over tables and text.
\begin{table}[t]
\begin{center}
\small
\begin{tabular}{llcc}
\toprule
\multicolumn{1}{c}{\bf Model} &
\multicolumn{1}{c}{\bf Setting} &
\multicolumn{1}{c}{\bf EM} &
\multicolumn{1}{c}{\bf F1} \\
\midrule

\multicolumn{4}{l}{\textbf{Task-specific and prior LLM baselines}} \\
TAGOP & -- & 47.05 & 47.60 \\
TAT-LLM (LoRA) & -- & 80.26 & 80.89 \\
TAT-LLM (FFT) & -- & 82.67 & 83.07 \\

\midrule
\multicolumn{4}{l}{\textbf{Reference models}} \\
Qwen2.5-7B-Instruct (Student) & Without tuning & 46.33 & 46.75 \\
Qwen2.5-72B-Instruct (Teacher, no hints) & Without tuning & 69.55 & 70.02 \\
Qwen2.5-72B-Instruct (Teacher, with hints) & Without tuning & 78.46 & 78.73 \\

\midrule
\multicolumn{4}{l}{\textbf{Our method}} \\
Qwen2.5-7B-Instruct (Attention-only) & Round 2 & 85.08 & 85.29 \\
Qwen2.5-7B-Instruct (Attention+MLP) & Round 2 & \textbf{87.00} & \textbf{87.18} \\

\bottomrule
\end{tabular}
\end{center}
\caption{Main results on the test set of the filtered TAT-QA subset. We compare against task-specific baselines, prior LLM-based methods, and reference teacher/student models.}
\label{tab:main_results}
\end{table}

\begin{table}[t]
\centering
\small
\begin{tabular}{l lcc cc}
\toprule
\textbf{Model} & \textbf{Round} & \multicolumn{2}{c}{\textbf{Dev}} & \multicolumn{2}{c}{\textbf{Test}} \\
\cmidrule(lr){3-4} \cmidrule(lr){5-6}
 &  & EM & F1 & EM & F1 \\
\midrule
Attention-only & Round 1 & 82.73 & 82.78 & 84.12 & 84.27 \\
Attention+MLP & Round 1 & 84.14 & 84.22 & 84.72 & 84.80 \\
Attention-only & Round 2 & 85.13 & 85.18 & 85.08 & 85.29 \\
Attention+MLP & Round 2 & \textbf{86.33} & \textbf{86.38} & \textbf{87.00} & \textbf{87.18} \\
\bottomrule
\end{tabular}
\caption{Performance across training rounds and adaptation settings on dev and test sets.}
\label{tab:training_analysis}
\end{table}
\textbf{Phase 1 Results}: The final teacher prompts, with and without derivation hints, were obtained through manual refinement and are shown in Fig.\ref{fig:prompt} in the appendix. After filtering, 5,190 of 6,603 training examples are retained. As shown in Table \ref{tab:training_analysis}, the resulting Round-1 student models achieve strong performance on both the development and test sets, with attention+MLP consistently outperforming attention-only. On the test set, the attention+MLP model achieves 84.72 EM and 84.80 F1, compared with 84.12 EM and 84.27 F1 for the attention-only model. Notably, both Round-1 student models already outperform all baselines and reference models reported in Table \ref{tab:main_results}.

\textbf{Phase 2 Results}: On the teacher-failed subset, the attention-only model recovers 781 additional examples, while the attention+MLP model recovers 979. As shown in Table~\ref{tab:training_analysis}, the second round of training improves performance over Round 1 for both variants on both the dev and test sets. For attention-only, performance improves from 82.73 EM / 82.78 F1 to 85.13 EM / 85.18 F1 on the dev set, and from 84.12 EM / 84.27 F1 to 85.08 EM / 85.29 F1 on the test set. For attention+MLP, performance improves from 84.14 EM / 84.22 F1 to 86.33 EM / 86.38 F1 on the dev set, and from 84.72 EM / 84.80 F1 to 87.00 EM / 87.18 F1 on the test set.

\section{Analysis}
\subsection{Performance Improvements from Different Techniques}
We analyze the drivers of the student model’s performance improvements by examining three factors: teacher-side data generation, LoRA module selection, and recursive training.

\begin{figure}[t]
\begin{center}
\includegraphics[width=0.8\linewidth]{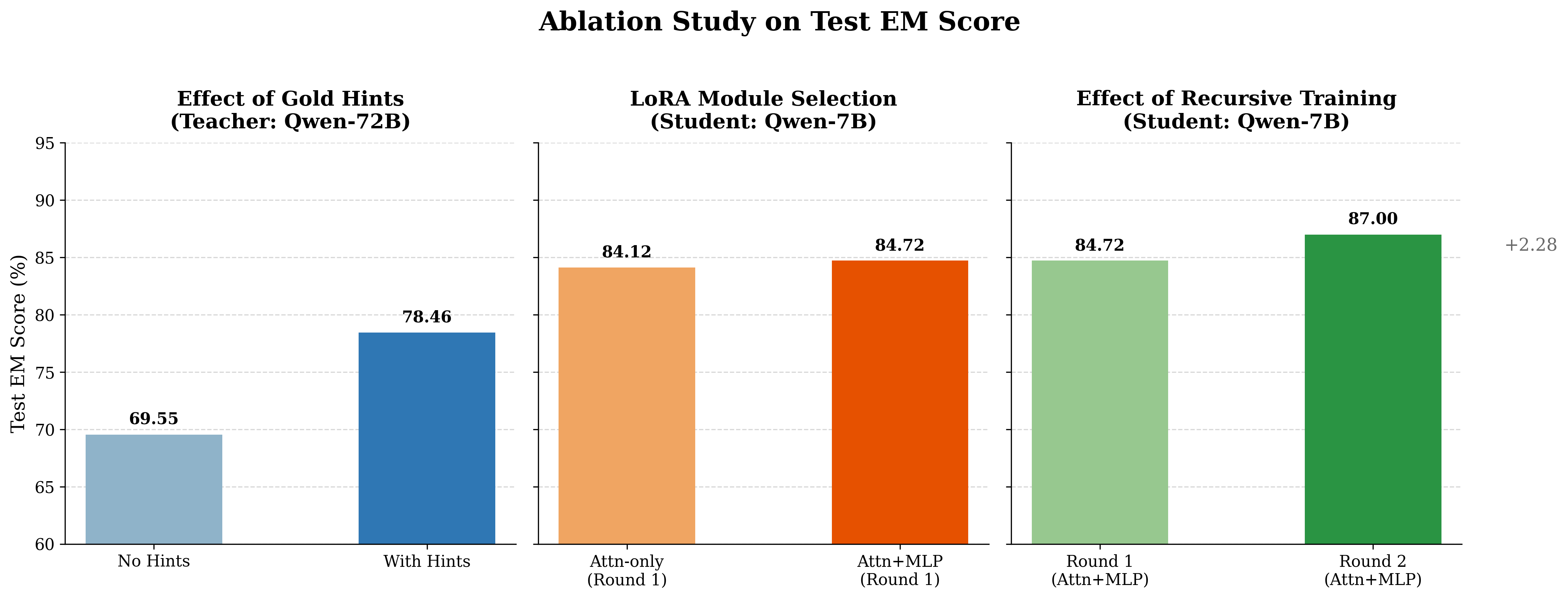} 
\end{center}
\caption{Ablation analysis evaluating the impact of core framework components on test EM scores.}
\label{fig:effect}
\end{figure}

First, we found that teacher-side data quality is critical for effective distillation. As shown in Fig. \ref{fig:effect}, providing ground-truth derivations as reasoning hints improves the 72B teacher’s test EM from 69.55 to 78.46. In addition, through manual and DSPy-based prompt refinement, we found that enforcing a structured output format substantially improved the executability and consistency of teacher-generated programs. Combined with strict execution-based filtering, these steps retain 78.6\% of the training examples and produce a reliable supervision set that directly enables strong Round-1 student performance.

Second, we evaluated the effect of expanding the Low-Rank Adaptation (LoRA) modules. Adapting both the attention and MLP projections consistently outperforms adapting the attention weights alone. In the initial fine-tuning phase, the attention+MLP student achieves a test EM of 84.72 and F1 of 84.80, surpassing the attention-only variant (84.12 EM / 84.27 F1). This structural advantage is maintained throughout the subsequent training phase.

Finally, we observe additional gains from our recursive training strategy (Phase 2). By recovering teacher-failed examples via student sampling and executing a second round of fine-tuning, performance improves for both architectural variants. The attention+MLP model reaches a peak test EM of 87.00 and F1 of 87.18. Phase 2 consistently improves over Round 1, although the size of the improvement varies across training configurations.
\subsection{Subgroup Performance}
\begin{figure}[t]
\begin{center}
\includegraphics[width=0.9\linewidth]{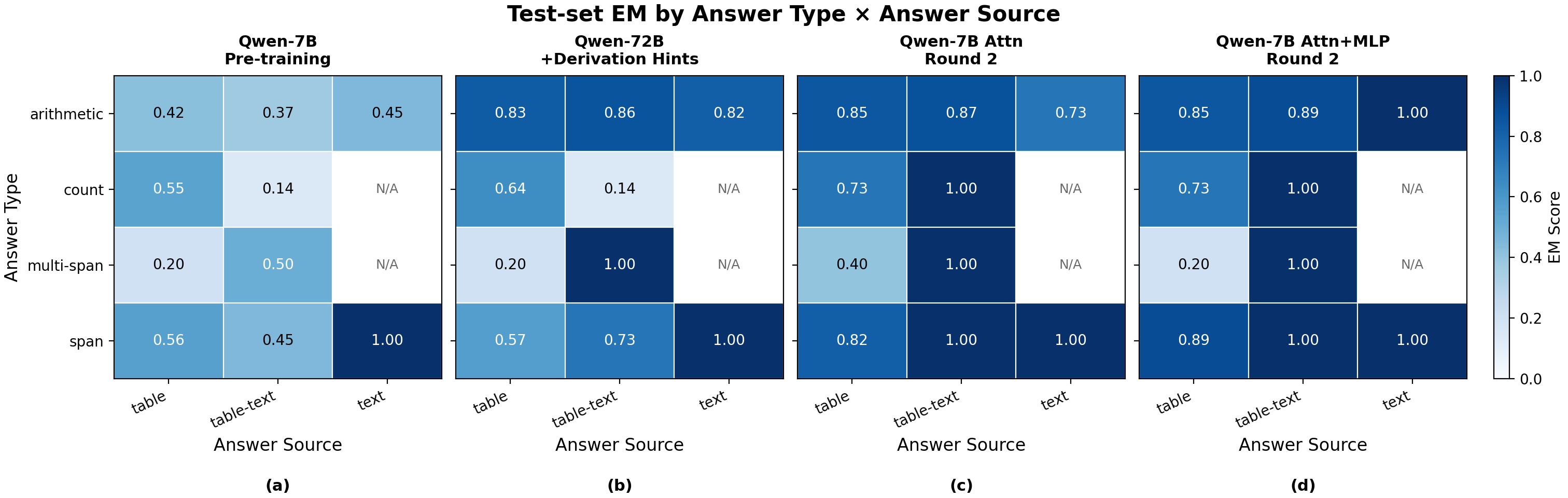} 
\end{center}
\caption{Test-set Exact Match (EM) scores by answer type and answer source across model stages: (a) Qwen-7B pre-training, (b) Qwen-72B with derivation hints, (c) Qwen-7B Round 2 (attention-only), and (d) Qwen-7B Round 2 (attention+MLP). Cells marked as N/A indicate that no examples are available for that subgroup.}
\label{fig:performace_type}
\end{figure}

Fig. \ref{fig:performace_type} shows test-set EM scores by answer type and answer source across different stages of our framework. The largest gains occur in categories requiring structured numerical reasoning over tables and mixed evidence. The student achieves strong EM across all question types in table-text settings, reaching perfect EM on count, multi-span, and span queries.
\\
The breakdown also shows large improvements on arithmetic questions, one of the hardest subsets of financial QA. Compared with the pretrained 7B model, programmatic distillation pushes the student above 85\% EM on both table and table-text cases, highlighting the value of delegating computation to an external interpreter. In addition, Round-2 self-improvement enables the student to surpass the teacher in several mixed-evidence settings. Finally, extending LoRA adaptation from attention-only to attention+MLP yields further gains, notably improving pure text arithmetic from 0.73 to 1.00.

\subsection{Error Composition}
\begin{figure}[t]
\begin{center}
\includegraphics[width=0.8\linewidth]{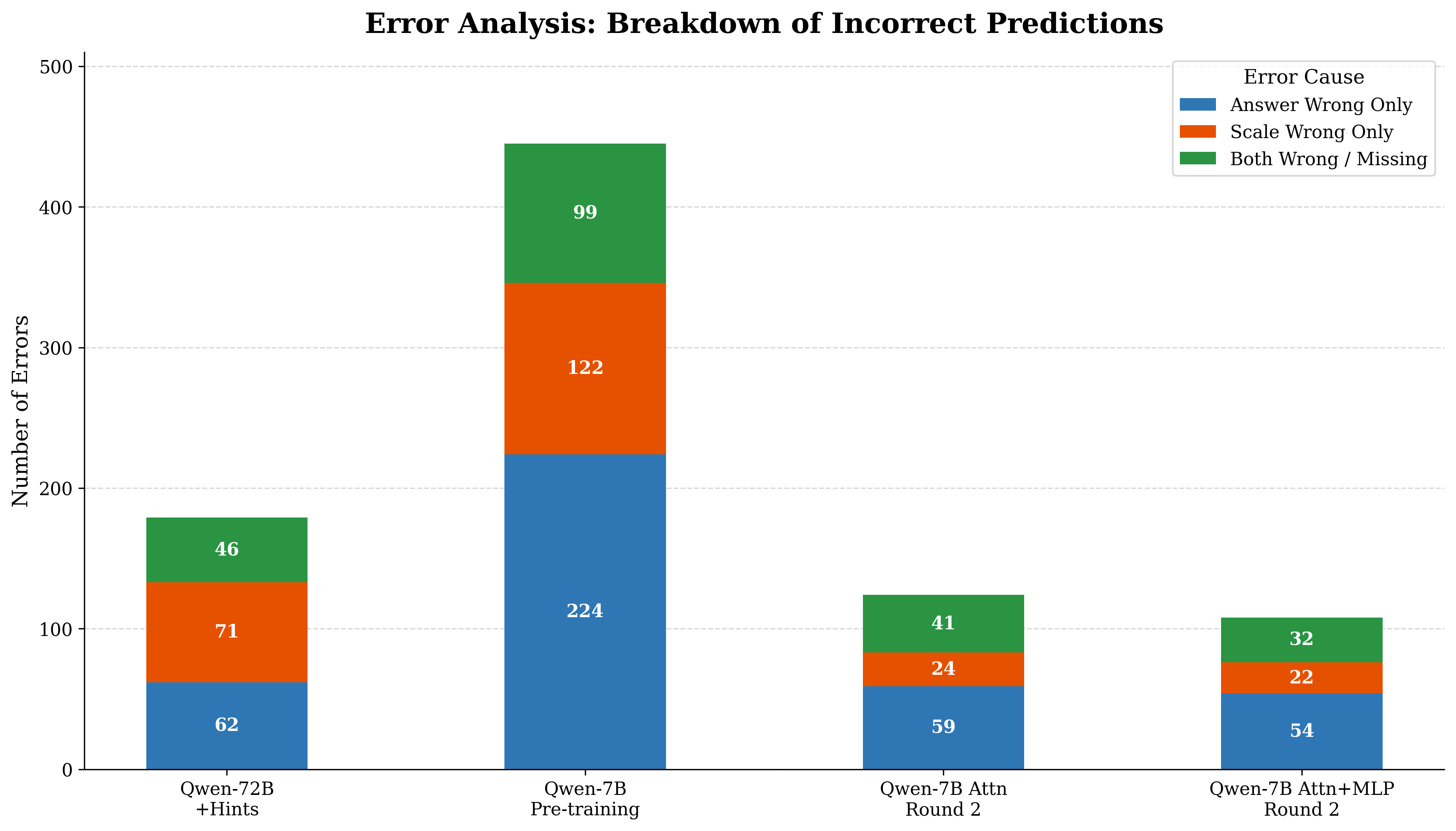} 
\end{center}
\caption{Breakdown of incorrect predictions with three categories: (1) answer wrong only, (2) scale wrong only, and (3) both wrong/missing.}
\label{fig:error_analysis}
\end{figure}

The error decomposition (Fig. ~\ref{fig:error_analysis}) shows that most of the remaining mistakes of the fine-tuned student arise from answer prediction errors rather than scale-only errors. Compared with the untuned Qwen-7B baseline, Round-2 training reduces the total number of errors from 445 to 124 for the attention-only model and 108 for the attention+MLP model.
\\
All three error categories decrease substantially after training: answer-wrong-only errors drop from 224 to 59 and 54, scale-wrong-only errors drop from 122 to 24 and 22, and both-wrong/missing cases drop from 99 to 41 and 32.\\
This suggests that programmatic distillation improves both reasoning correctness and output-format consistency. Overall, compared with the untuned Qwen-7B baseline, both fine-tuned Qwen-7B models substantially reduce all three error types, and even outperform the Qwen-72B teacher model.
\section{Limitations}
Although our approach yields strong improvements overall, several limitations remain. First, our evaluation uses a filtered subset of TAT-QA designed for executable program generation. As a result, the evaluation set is relatively small, and some answer type-source combinations are absent, especially simpler extraction-based cases that do not require derivation. Second, despite strong gains, all three error types are not yet fully resolved, indicating that answer prediction, scale prediction, and joint failures still remain.
\section{Conclusion and Future Work}
We propose Gold-Guided Programmatic Distillation, a framework that improves the numerical reasoning capabilities of compact language models through execution-verified programs and iterative self-improvement. On TAT-QA, our 7B student achieves 87.00 EM and 87.18 F1, substantially outperforming both the TAGOP baseline and the 72B teacher with derivation hints. Our results show that this framework significantly improves arithmetic reliability and reduces numerical and scale-related errors, with the largest gains on tasks requiring reasoning over mixed tabular and textual evidence.
\\
Despite these improvements, the problem is not fully resolved. All three types of errors (answer prediction errors, scale errors, and joint failures) remain present after training, indicating that both reasoning accuracy and output alignment can be further improved.
\\
For future work, a promising direction is to incorporate a task-aware, agent-like routing mechanism that determines whether a query requires programmatic reasoning or simpler extraction, and selects the appropriate solving strategy accordingly. This could involve different prompts, solvers, or model components, rather than relying on a single unified setup. In addition, evaluating this framework on broader financial QA benchmarks, such as FinQA and ConvFinQA, will be important for assessing its generalizability.
\bibliography{colm2026_conference}

@inproceedings{zhu2021tat,
  title={TAT-QA: A question answering benchmark on a hybrid of tabular and textual content in finance},
  author={Zhu, Fengbin and Lei, Wenqiang and Huang, Youcheng and Wang, Chao and Zhang, Shuo and Lv, Jiancheng and Feng, Fuli and Chua, Tat-Seng},
  booktitle={Proceedings of the 59th annual meeting of the Association for Computational Linguistics and the 11th international joint conference on natural language processing (volume 1: long papers)},
  pages={3277--3287},
  year={2021}
}

@inproceedings{rajpurkar2016squad,
  title={Squad: 100,000+ questions for machine comprehension of text},
  author={Rajpurkar, Pranav and Zhang, Jian and Lopyrev, Konstantin and Liang, Percy},
  booktitle={Proceedings of the 2016 conference on empirical methods in natural language processing},
  pages={2383--2392},
  year={2016}
}

@article{kuhn1955hungarian,
  title={The Hungarian method for the assignment problem},
  author={Kuhn, Harold W},
  journal={Naval research logistics quarterly},
  volume={2},
  number={1-2},
  pages={83--97},
  year={1955},
  publisher={Wiley Online Library}
}

@inproceedings{zhu2024tat,
  title={Tat-llm: A specialized language model for discrete reasoning over financial tabular and textual data},
  author={Zhu, Fengbin and Liu, Ziyang and Feng, Fuli and Wang, Chao and Li, Moxin and Chua, Tat Seng},
  booktitle={Proceedings of the 5th ACM International Conference on AI in Finance},
  pages={310--318},
  year={2024}
}

@article{wei2022chain,
  title={Chain-of-thought prompting elicits reasoning in large language models},
  author={Wei, Jason and Wang, Xuezhi and Schuurmans, Dale and Bosma, Maarten and Xia, Fei and Chi, Ed and Le, Quoc V and Zhou, Denny and others},
  journal={Advances in neural information processing systems},
  volume={35},
  pages={24824--24837},
  year={2022}
}

@inproceedings{gao2023pal,
  title={Pal: Program-aided language models},
  author={Gao, Luyu and Madaan, Aman and Zhou, Shuyan and Alon, Uri and Liu, Pengfei and Yang, Yiming and Callan, Jamie and Neubig, Graham},
  booktitle={International conference on machine learning},
  pages={10764--10799},
  year={2023},
  organization={PMLR}
}

@article{chen2022program,
  title={Program of thoughts prompting: Disentangling computation from reasoning for numerical reasoning tasks},
  author={Chen, Wenhu and Ma, Xueguang and Wang, Xinyi and Cohen, William W},
  journal={arXiv preprint arXiv:2211.12588},
  year={2022}
}

@inproceedings{magister2023teaching,
  title={Teaching small language models to reason},
  author={Magister, Lucie Charlotte and Mallinson, Jonathan and Adamek, Jakub and Malmi, Eric and Severyn, Aliaksei},
  booktitle={Proceedings of the 61st Annual Meeting of the Association for Computational Linguistics (Volume 2: Short Papers)},
  pages={1773--1781},
  year={2023}
}

@inproceedings{hsieh2023distilling,
  title={Distilling step-by-step! outperforming larger language models with less training data and smaller model sizes},
  author={Hsieh, Cheng-Yu and Li, Chun-Liang and Yeh, Chih-Kuan and Nakhost, Hootan and Fujii, Yasuhisa and Ratner, Alex and Krishna, Ranjay and Lee, Chen-Yu and Pfister, Tomas},
  booktitle={Findings of the Association for Computational Linguistics: ACL 2023},
  pages={8003--8017},
  year={2023}
}

@inproceedings{zhu2024pad,
  title={PaD: Program-aided distillation can teach small models reasoning better than chain-of-thought fine-tuning},
  author={Zhu, Xuekai and Qi, Biqing and Zhang, Kaiyan and Long, Xinwei and Lin, Zhouhan and Zhou, Bowen},
  booktitle={Proceedings of the 2024 Conference of the North American Chapter of the Association for Computational Linguistics: Human Language Technologies (Volume 1: Long Papers)},
  pages={2571--2597},
  year={2024}
}

@article{zelikman2022star,
  title={Star: Bootstrapping reasoning with reasoning},
  author={Zelikman, Eric and Wu, Yuhuai and Mu, Jesse and Goodman, Noah},
  journal={Advances in Neural Information Processing Systems},
  volume={35},
  pages={15476--15488},
  year={2022}
}

@article{yuan2023scaling,
  title={Scaling relationship on learning mathematical reasoning with large language models},
  author={Yuan, Zheng and Yuan, Hongyi and Li, Chengpeng and Dong, Guanting and Lu, Keming and Tan, Chuanqi and Zhou, Chang and Zhou, Jingren},
  journal={arXiv preprint arXiv:2308.01825},
  year={2023}
}

@article{tian2025not,
  title={Not all correct answers are equal: Why your distillation source matters},
  author={Tian, Xiaoyu and Ji, Yunjie and Wang, Haotian and Chen, Shuaiting and Zhao, Sitong and Peng, Yiping and Zhao, Han and Li, Xiangang},
  journal={arXiv preprint arXiv:2505.14464},
  year={2025}
}

@article{bai2023qwen,
  title={Qwen technical report},
  author={Bai, Jinze and Bai, Shuai and Chu, Yunfei and Cui, Zeyu and Dang, Kai and Deng, Xiaodong and Fan, Yang and Ge, Wenbin and Han, Yu and Huang, Fei and others},
  journal={arXiv preprint arXiv:2309.16609},
  year={2023}
}

@article{hu2022lora,
  title={Lora: Low-rank adaptation of large language models.},
  author={Hu, Edward J and Shen, Yelong and Wallis, Phillip and Allen-Zhu, Zeyuan and Li, Yuanzhi and Wang, Shean and Wang, Liang and Chen, Weizhu and others},
  journal={Iclr},
  volume={1},
  number={2},
  pages={3},
  year={2022}
}

@misc{khattab2023dspycompilingdeclarativelanguage,
      title={DSPy: Compiling Declarative Language Model Calls into Self-Improving Pipelines}, 
      author={Omar Khattab and Arnav Singhvi and Paridhi Maheshwari and Zhiyuan Zhang and Keshav Santhanam and Sri Vardhamanan and Saiful Haq and Ashutosh Sharma and Thomas T. Joshi and Hanna Moazam and Heather Miller and Matei Zaharia and Christopher Potts},
      year={2023},
      eprint={2310.03714},
      archivePrefix={arXiv},
      primaryClass={cs.CL},
      url={https://arxiv.org/abs/2310.03714}, 
}

@misc{peft,
  title        = {PEFT: Parameter-Efficient Fine-Tuning},
  author       = {{Hugging Face}},
  year         = {2023},
  howpublished = {\url{https://github.com/huggingface/peft}},
  note         = {GitHub repository}
}
\bibliographystyle{colm2026_conference}
\clearpage
\section{Appendix}
\subsection{Prompt}
\begin{figure}[h]
\begin{center}
\includegraphics[width=\linewidth]{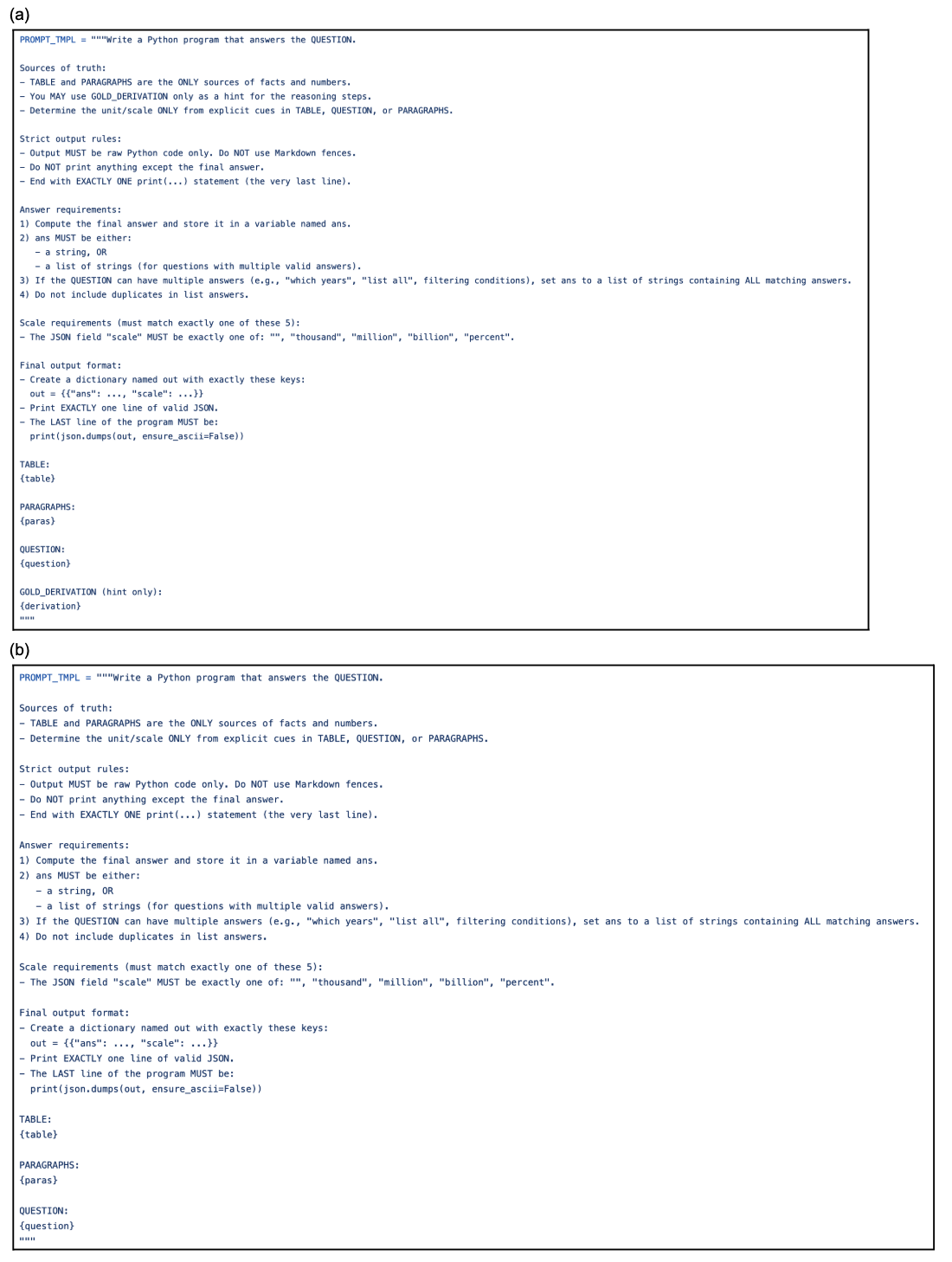}
\end{center}
\caption{Prompts used in our pipeline. (a) Teacher prompt with derivation hints. (b) Shared prompt without derivation hints for teacher and student.}
\label{fig:prompt}
\end{figure}
\clearpage
\subsection{Examples}
\begin{figure}[h]
\begin{center}
\includegraphics[width=\linewidth]{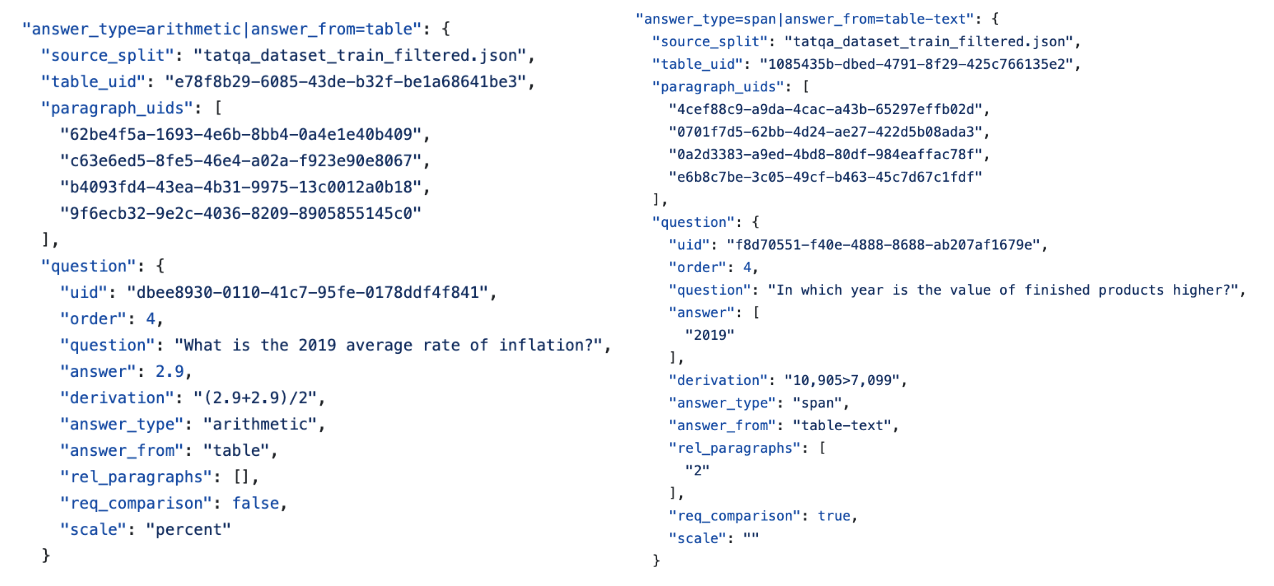}
\end{center}
\caption{Examples of TAT-QA dataset.}
\label{fig:example}
\end{figure}

\end{document}